\RequirePackage[]{graphicx} % Loaded here to show images in draft mode, too
\graphicspath{{./Figures/}{./Graphs/}{./Drawings/}}

\documentclass[]{arxsub}

\usepackage[utf8]{inputenc}
\usepackage{graphicx}
\usepackage{environ}
\usepackage{listings}
\usepackage{booktabs}
\usepackage{blindtext}
\usepackage{url}

\DeclareMathOperator*{\argmin}{arg\,min}

\newcommand{\figref}[2][]{\hyperref[#2]{\mbox{Figure \ref*{#2}#1}}}
\newcommand{\tblref}[2][]{\hyperref[#2]{\mbox{Table \ref*{#2}#1}}}
\renewcommand{\eqref}[1]{\hyperref[#1]{\mbox{Eq. \ref*{#1}}}}

\newcommand{\norm}[1]{\left\lVert#1\right\rVert}
\newcommand{\id}{\mathbb{1}}

\providecommand{\doi}{}
\renewcommand{\doi}[1]{\href{https://doi.org/#1}{#1}}

\newcommand{\preSubmission}{Part of this work has been presented at the ISMRM Annual Conference 2021 (Virtual).}

\providecommand{\del}[2][]{}

\providecommand{\dels}[2][]{}

\providecommand{\printfunding}{}

\title{Deep, Deep Learning with BART}

\corres{Martin Uecker, Graz University of Technology,
Institute of Biomedical Imaging, Stremayrgasse~16/3, 8010~Graz, AUSTRIA, \email{uecker@tugraz.at}}

\author[1]{Moritz Blumenthal}{}
\author[1]{Guanxiong Luo}{}
\author[1]{Martin Schilling}{}
\author[2]{H. Christian M. Holme}{}
\author[1,2,3,4,5]{Martin Uecker}{}

\jnlcitation{\cname{%
\author{M. Blumenthal}, 
\author{G. Luo}, 
\author{M. Schilling}, 
\author{H. C. M. Holme}, and 
\author{M. Uecker}} (\cyear{2022}), 
\ctitle{Deep, Deep Learning with BART}, \cjournal{}, \cvol{}.
}

\authormark{M. Blumenthal \textsc{et al.}}

\address[1]{Institute for Diagnostic and Interventional Radiology, University Medical Center Göttingen, Göttingen, Germany}
\address[2]{Institute of Biomedical Imaging, Graz University of Technology, Graz, Austria}
\address[3]{German Centre for Cardiovascular Research (DZHK),Partner Site Göttingen, Göttingen, Germany}
\address[4]{Cluster of Excellence ``Multiscale Bioimaging: from Molecular Machines to Networks of Excitable Cells'' (MBExC), University of G\"ottingen, Germany}
\address[5]{BioTechMed-Graz, Graz, Austria}

\articletype{Research Article}

\finfo{
This work was funded by the "Niedersächsisches Vorab" funding line of the Volkswagen Foundation,
and funded in part by NIH under grant U24EB029240
and funded in part by the Deutsche Forschungsgemeinschaft (DFG, German Research Foundation) under grant UE 189/1-1, under Germany’s Excellence Strategy - EXC 2067/1- 390729940, and with Project-ID 432680300 - SFB 1456.
This work was supported by the DZHK (German Centre for Cardiovascular Research).
}

\abstract{
\section{Purpose}
To develop a deep-learning-based image reconstruction framework
for reproducible research in MRI.
\section{Methods} The BART toolbox offers a rich set of implementations of calibration and
reconstruction algorithms for parallel imaging and compressed sensing.
In this work, BART was extended by a non-linear operator framework that provides
automatic differentiation to allow computation of gradients.
Existing MRI-specific operators of BART, such as the non-uniform fast Fourier transform, are
directly integrated into this framework and are complemented by common building blocks
used in neural networks.
To evaluate the use of the framework for advanced deep-learning-based reconstruction,
two state-of-the-art unrolled reconstruction networks, namely the
Variational Network \cite{Hammernik_Magn.Reson.Med._2017} and MoDL \cite{Aggarwal_IEEETrans.Med.Imag._2019},
were implemented.
\section{Results}
State-of-the-art deep image-reconstruction networks can be constructed
and trained using BART's gradient based optimization algorithms. The BART
implementation achieves a similar performance in terms of training time
and reconstruction quality compared to the original implementations
based on TensorFlow.
\section{Conclusion}
By integrating non-linear operators and neural networks into BART, we provide
a general framework for deep-learning-based reconstruction in MRI.
}

\begin{document}

\maketitle

\section{Introduction}

In the last decades, magnetic resonance imaging (MRI) has advanced substantially in terms of acquisition speed and image quality.
Parallel imaging utilizes the signal of multiple receiver coils for image reconstruction
by combining the signals in k-space \cite{Sodickson_Magn.Reson.Med._1997, Griswold_Magn.Reson.Med._2002, Lustig_Magn.Reson.Med._2010}
or image space \cite{Pruessmann_Magn.Reson.Med._1999}.
Another step towards the current state-of-the-art image reconstruction was
the use of compressed sensing for MRI \cite{Lustig_Magn.Reson.Med._2007, Block_Magn.Reson.Med._2007}.
Advanced methods now integrate compressed sensing and parallel imaging by using
sparsifying regularization terms when solving the inverse problem
for parallel imaging \cite{Block_Magn.Reson.Med._2007, Liang_Magn.Reson.Med._2009}.
These techniques admit a Bayesian interpretation where regularization
terms can be understood as the integration of prior knowledge into the reconstruction.

In recent years, deep learning has become a major research interest in image reconstruction
with the goal to improve upon the previously used hand-crafted regularization terms by
learning image properties from large data sets.
The public availability of deep learning frameworks such as TensorFlow \cite{Abadi__2016}
or PyTorch \cite{Paszke__2019} simplifies access to deep learning methods for MRI researchers.
Moreover, public data sets from \url{mridata.org} \cite{Sawyer__2013, Ong__2018}
and from the fastMRI challenge \cite{Knoll_Magn.Reson.Med._2020} provide
a large amount of training data and open the field of research to data
scientists not having access to MRI data.

Neural networks have been utilized in various ways for MRI reconstruction.
Some authors have proposed to learn a direct mapping from the acquired k-space data to the image domain \cite{Zhu_Nature_2018}.
However, these methods usually lack a data-consistency guarantee, i.e. the output of the reconstruction may not be consistent with the measured k-space data.
Others have used neural network to train regularizers which can be used for image reconstruction in a subsequent step \cite{Luo_Magn.Reson.Med._2020}.
In one class of such regularizers, a neural network is trained to enhance an initial reconstruction. Afterwards, the $\ell_2$ difference to this reconstruction is used as regularizer \cite{Yang_IEEETrans.Med.Imag._2018, Kofler_Phys.Med.Biol._2020}.
Another class of networks with data consistency are networks that model an unrolled iterative optimization algorithm \cite{Hammernik_Magn.Reson.Med._2017, Aggarwal_IEEETrans.Med.Imag._2019, Schlemper_IEEETrans.Med.Imag._2018}.
In each iteration of such a network, the network part, usually a CNN or U-Net, updates the current reconstruction and afterwards soft data consistency is imposed by a gradient step or proximal mapping.
The resulting unrolled networks are then trained as an end-to-end mapping from the k-space to the image domain.

BART \cite{Uecker__2015} is an open-source framework providing implementations of various calibration methods
and reconstruction algorithms for parallel imaging and compressed sensing.
It consists of programming libraries and command line tools for easy but flexible access to the programming libraries.
BART is developed with the purpose of facilitating reproducible research and has a focus on backwards compatibility,
while still offering rapid prototyping and testing of advanced reconstruction algorithm with the goal of translating them into clinical reconstruction pipelines.
The high-level reconstruction algorithms of BART are built around programming libraries offering generic implementations of various iterative algorithms as well as an efficient numerical backend.
The backend provides functions acting on multidimensional arrays (or tensors) which support acceleration by multiple threads or (multiple) graphical processing units (GPUs).
In this work, we extend BART with a complete framework for non-linear operators.
The framework builds on our previous work on non-linear calibrationless parallel imaging \cite{Holme_Sci.Rep._2019}
and physics-based reconstruction \cite{Wang_Philos.Trans.R.Soc.A._2021}, and is now extended with automatic differentiation,
additional building blocks for neural networks, and new optimization algorithms \cite{Blumenthal__2021}.
In combination with the powerful numerical backend, the non-linear operator framework can then be used to efficiently train neural networks.
Moreover, non-linear operators can be used to wrap around TensorFlow graphs, allowing the integration of pre-trained networks into BART's reconstruction algorithms \cite{Luo__2021a}. 
MRI reconstruction networks imposing data consistency require a large amount of domain specific knowledge.
By integrating neural networks into BART, we benefit from BART's rich set of MRI specific modules and algorithms which
can be easily reused for deep learning based MRI reconstructions.
Written in C and only depending on a few external libraries, we consider BART a solid basis for future
research that integrates classical image reconstruction with deep learning.

In the remainder of this manuscript we first describe in detail the implementation of our deep learning framework and its integration into BART.
There, we focus on the numerical backend, the automatic differentiation, the iterative training algorithms and the neural network framework.
Afterwards, we present our implementation of the Variational Network (VarNet)\cite{Hammernik_Magn.Reson.Med._2017} and MoDL \cite{Aggarwal_IEEETrans.Med.Imag._2019}, and compare their performance to the original implementations based on TensorFlow.

  %newlines are needed

\section{Methods}
\label{sec:methods}

A neural network is a non-linear function $F$ mapping the input data $\mathbf{x}$ and weights $\mathbf{\theta}$ to an output $\mathbf{y}=F(\mathbf{x};\mathbf{\theta})$.
Training a neural network corresponds to fitting the neural network to a training dataset by minimizing some suitable loss $L$, i.e.
\begin{equation}
    \mathbf{\theta}^*=\argmin_{\mathbf{\theta}} \left[\sum_i L(\mathbf{y}_i, F(\mathbf{x}_i;\mathbf{\theta}))\right] \label{eq:network_optimize}\,.
\end{equation}
Usually, neural networks are constructed from small building blocks such as fully-connected layers, convolutional layers or activation functions.
Automatic differentiation is used to compute the gradients of the loss needed for gradient-based optimization algorithms such as stochastic gradient descent or ADAM \cite{1412.6980}.
Offering automatic differentiation and efficient implementations for the small building blocks are key features of deep learning frameworks.
In the first part of this section, we describe the integration of programming libraries used for deep learning in BART, before we describe our implementations of VarNet and MoDL in the second part.

\subsection{Libraries for Deep Learning in BART}

\begin{figure}
    \centering
    \includegraphics[width=.7\linewidth]{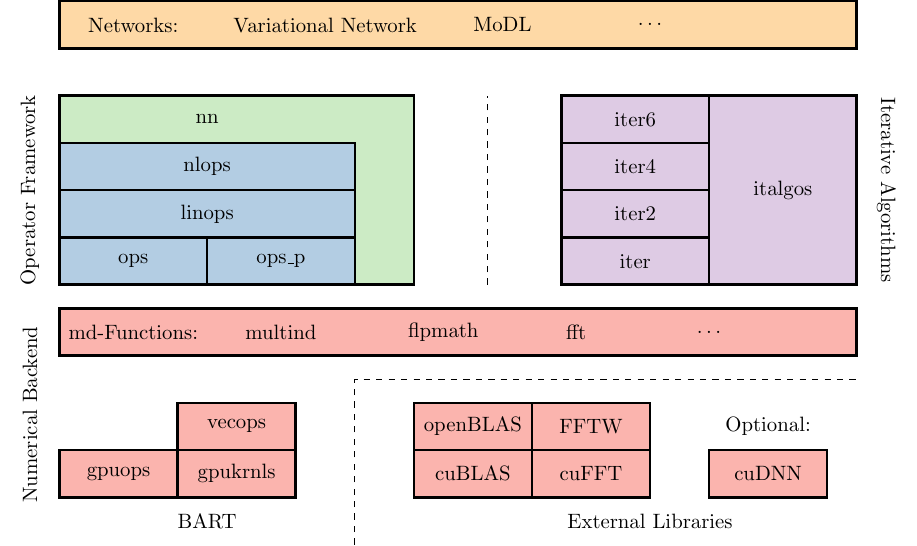}
    \caption{
        Integration of deep learning modules into BART.
        The numerical backend (red) is accessed by md-functions which invoke BART's internal generically-optimized
        functions or external libraries offering highly optimized code for special functions.
        Differentiable neural networks are implemented as non-linear operators (blue).
        The nn-library (green) extends the non-linear operator framework by deep learning specific features.
        The training algorithms are integrated in BART's iterative framework (violet).
        Iter6 provides a new interface for batched gradient-based training algorithms.
    }
    \label{fig:integration}
\end{figure}

The basic integration of libraries used for neural networks in BART is depicted in \figref{fig:integration}.
The backend provide access to optimized numerical functions.
Based on the backend, the non-linear operator framework is used to construct neural networks from small building blocks and provides automatic differentiation to compute gradients.
The nn-library then extends this non-linear operator framework by deep learning specific functions.
Finally, new training algorithms for deep learning are integrated in BART's iterative framework.

\subsubsection{Numerical Backend}

The numerical backend of BART is designed around functions acting on multidimensional (md) arrays.
An md-array is described by its dimensions $\mathbf{d}\in \mathbb{N}^N$, its rank $N$ and, optionally, its strides $\mathbf{s}\in \mathbb{Z}^N$ describing how an element of the md-array is accessed in memory.
The offset of an element at position $\mathbf{p}\in \mathbb{N}^N$ is given by $o = \mathbf{p}\cdot\mathbf{s}$.
By default, BART assumes column-major ordering, i.e. the first dimension of the md-array is stored continuously in memory corresponding to the strides $\mathbf{s}_i = \prod^{i-1}_{j=0} \mathbf{d_j}$.
By manipulating the strides, different views on the memory can be generated without copying data.
The memory for an md-array can be allocated on the CPU or on the GPU.
On supported GPUs, the GPU memory can be oversubscribed, i.e. GPU memory is automatically swapped by the driver to CPU memory.

\paragraph{Md-Functions}

Md-functions provide a consistent and flexible interface to functions acting on md-arrays.
They loop over all positions defined by the dimensions and apply a scalar-valued kernel on the elements accessed using the provided strides.
For example, \texttt{md\_fmac2} applies
\begin{align}
\text{for } \mathbf{p} \in \{0, ..., \mathbf{d}_0-1\}\times\dots\times \{0, ..., \mathbf{d}_{N-1} - 1\}:\notag\\
a[\mathbf{p}\cdot\mathbf{s}^a] \leftarrow a[\mathbf{p}\cdot\mathbf{s}^a] + b[\mathbf{p}\cdot\mathbf{s}^b] \cdot c[\mathbf{p}\cdot\mathbf{s}^c]\,.
\end{align}
By setting the strides correspondingly, many functions such as convolutions, matrix-vector multiplications or a dot product can be derived.
For example, if $\mathbf{s}^a=0$, the dot product of $b$ and $c$ is accumulated in $a[0]$.
If the memory of an md-array is located on the GPU, the computation of a md-function is automatically executed on the GPU.
The loops of the md-functions are generically optimized in the backend.
Further, strides corresponding to specific operations such as matrix-matrix multiplication or convolution are detected, and specialized code -- possibly from external libraries such as cuBLAS or cuDNN -- is executed. Thus, md-functions provide a generic but still efficient interface to the numeric backend.

\paragraph{Bitwise Reproducibility}

Floating point arithmetic is not associative making multi-threaded programs non-deterministic if the order of the operations depends
on the runtime of individual threads.
BART's GPU kernels and the generic parallelization do not introduce any non-deterministic operations except for the gridding code 
of the nuFFT.
cuBLAS and cuDNN are deterministic across runs when executed on GPUs with the same architecture except for some specific functions.
By default, BART makes only use of these deterministic functions, however, the compile-time option \texttt{NON\_DETERMINISTIC=1} can
be used to allow BART to select non-deterministic algorithms to improve computational performance.

\subsubsection{Automatic Differentiation and the Non-Linear Operator Framework}  Usually, neural networks are trained by gradient-based methods.
The vanilla version of a gradient descent algorithm optimizes \eqref{eq:network_optimize} by the iteration $\mathbf{\theta}_{i+1} = \mathbf{\theta}_{i}-\eta\nabla_{\mathbf{\theta}} \left(\sum_i L(\mathbf{y}_i, F(\mathbf{x}_i;\mathbf{\theta})\right)$, where $\eta$ is the learning rate and $\nabla_{\mathbf{\theta}}$ denotes the gradient of the loss with respect to the \mbox{weights $\mathbf{\theta}$}. Automatic differentiation, as described below, allows to construct an operator computing $\sum_i L(\mathbf{y}_i, F(\mathbf{x}_i;\mathbf{\theta})$ and to compute its derivative, i.e. gradient, with respect to the weights.

We first describe the automatic differentiation framework on an abstract level for real-valued operators, before we describe the extension to complex variables and the implementation details in subsequent paragraphs.
A non-linear operator (\texttt{nlop}) $F$ consists of the forward operator $F$ itself and its (Fréchet)-derivative $\mathrm{D} F|_{\mathbf{x}}$ - a linear operator (\texttt{linop}) applying the Jacobian matrix $J|_{\mathbf{x}}$ evaluated at some position $\mathbf{x}$ on its input -, i.e,
\begin{align}
    F:  &\mathbb{R}^N \to \mathbb{R}^M          & \mathrm{D} F|_{\mathbf{x}}: &\mathbb{R}^N \to \mathbb{R}^M       & J_{ij} &= \frac{\partial F_i}{\partial x_j}\notag\\
         &\mathbf{x} \mapsto \mathbf{y}=F(\mathbf{x})    &               &\mathrm{d}\mathbf{x} \mapsto \mathrm{d}\mathbf{y}=J|_{\mathbf{x}}\mathrm{d}\mathbf{x}\;.
\end{align}
Usually, the Jacobian $J$ is not stored explicitly, instead, the derivative $\mathrm{D}F|_{\mathbf{x}}$ or its transposed $\mathrm{D}F^T|_{\mathbf{x}}$ can be applied on test inputs.
By applying the derivative on a vector $\hat{e}_k$ containing zeros and a one at index $k$, the $k$-th column of the Jacobian is computed.
Correspondingly, the $k$-th row of $J$ is computed by applying the transposed derivative on $\hat{e}_k$.
In the special case $F:\mathbb{R}^N \to \mathbb{R}$ mapping to a scalar, the Jacobian reduces to a $1\times N$-matrix containing the gradient of $F$ which can be computed by applying $\mathrm{D}F|_{\mathbf{x}}^T$ on the scalar one, i.e.
\begin{align}
    \nabla F =
    \begin{pmatrix}
    \frac{\partial F_1}{\partial x_1}& \dots & \frac{\partial F_1}{\partial x_N}
    \end{pmatrix}^T
    =J^T=\mathrm{D}F^T\begin{pmatrix}1\end{pmatrix}\;.
    \label{eq:gradient_transposed}
\end{align}
Gradients are usually computed by the transposed derivative since this only requires one application of $\mathrm{D}F|_{\mathbf{x}}^T$ instead of $N$ applications of $\mathrm{D}F|_{\mathbf{x}}$ to compute each column of $J$ independently.
As depicted in \figref[A]{fig:nlops}, \texttt{nlop}s can have multiple inputs and outputs and there is a derivative for each combination of input and output. The derivatives are always evaluated at the inputs of the last call of $F$ and there is a shared data structure to communicate this information. For example, the multiplication operator $F(\mathbf{x}_1, \mathbf{x}_2)=\mathbf{x}_1\mathbf{x}_2$ stores $\mathbf{x}_2$ (and $\mathbf{x}_1$) needed by the derivative $\mathrm{D}_{\mathbf{x}_1} F|_{\mathbf{x}_1,\mathbf{x}_2}: \mathrm{d} \mathbf{x}_1 \mapsto \mathbf{x}_2\mathrm{d}\mathbf{x}_1$.

\paragraph{Composing Operators}
The crucial part of automatic differentiation is the possibility to \textit{chain} \texttt{nlop}s and compute the chained derivatives.
\figref[B]{fig:nlops} shows the \textit{chain} $H = G \circ F$ with its derivative $\mathrm{D}H|_{\mathbf{x}}=\mathrm{D}G|_{F(\mathbf{x})}\circ\mathrm{D}F|_{\mathbf{x}}$.
As $G$ is applied on $F(\mathbf{x})$, the derivative $\mathrm{D}G$ is automatically evaluated at $F(\mathbf{x})$.
To compute the transposed $\mathrm{D}H|_{\mathbf{x}}^T=\mathrm{D}F|_{\mathbf{x}}^T\circ\mathrm{D}G|_{F(\mathbf{x})}^T$, the $\mathrm{D}F^T$ and $\mathrm{D}G^T$ are applied in reverse order, hence the name backpropagation.
Similar to the \textit{chain}, BART provides a set of functions for composing \texttt{nlop}s with multiple inputs and outputs.
These functions can be used to \textit{combine} two \texttt{nlop}s to one, to \textit{link} an output of an \texttt{nlop} into one of its inputs, and to \textit{duplicate} one of its inputs into another one.
The action of these functions is presented in \figref[C]{fig:nlops}, where we demonstrate how two \texttt{nlop}s $F$ and $G$ can be used to construct an \texttt{nlop} computing $J(\mathbf{x}_1, \mathbf{x}_2)=F(\mathbf{x}_1, G(\mathbf{x}_1, \mathbf{x}_2))$. 
The resulting \texttt{nlop}s hold references to the base \texttt{nlop}s to call them, and the derivatives are constructed automatically.
Since \textit{combine}, \textit{duplicate} and \textit{link} can be nested, generic compositions of \texttt{nlop}s are possible.

\paragraph{Complex numbers}
\texttt{nlop}s in BART work with complex numbers, i.e. single precision complex floats.
The automatic differentiation framework is extended to complex numbers by identifying $\mathbb{C} \sim \mathbb{R}^2$. For example, a univariate complex mapping $F:x\mapsto y =F(x)$ is represented by
\begin{align}
    F: \begin{pmatrix}x_r\\ x_i\end{pmatrix} &\mapsto \begin{pmatrix}y_r\\y_i \end{pmatrix}
    &
    \mathrm{D}F: \begin{pmatrix}\mathrm{d}x_r\\ \mathrm{d}x_i\end{pmatrix} \mapsto 
    \begin{pmatrix}
        \frac{\partial y_r}{\partial x_r} & \frac{\partial y_r}{\partial x_i} \\
        \frac{\partial y_i}{\partial x_r} & \frac{\partial y_i}{\partial x_i}
    \end{pmatrix}
    \begin{pmatrix}\mathrm{d}x_r\\ \mathrm{d}x_i\end{pmatrix}\,.
\end{align}
This approach is equivalent to the so-called Wirtinger\cite{Wirtinger1927} or $\mathbb{{CR}}$-calculus\cite{KreutzDelgado_arXiv_2009} which introduces the complex derivatives $\frac{\partial y}{\partial x}$ and $\frac{\partial y}{\partial \bar{x}}$ to reformulate $\mathrm{D}F$ as 
\begin{align}
    \mathrm{D}F:\mathrm{d}x\mapsto&\tfrac{\partial y}{\partial x}\mathrm{d}x + \tfrac{\partial y}{\partial \bar{x}}\bar{\mathrm{d}x}&
    &\text{with}&
    \tfrac{\partial y}{\partial x}&=\tfrac{1}{2}\left(\tfrac{\partial y}{\partial x_r} - i \tfrac{\partial y}{\partial x_i}\right) 
    \notag\\
    \mathrm{D}F^T:\mathrm{d}y\mapsto&\bar{\tfrac{\partial y}{\partial x}} \mathrm{d}y + \tfrac{\partial y}{\partial \bar{x}} \mathrm{d}\bar{y}&
    &&
    \tfrac{\partial y}{\partial \bar{x}}&=\tfrac{1}{2}\left(\tfrac{\partial y}{\partial x_r} + i \tfrac{\partial y}{\partial x_i}\right)\,.
\end{align}
If $F$ is holomorphic, $\frac{\partial y}{\partial \bar{x}} = 0$ holds, such that $\mathrm{D}F$ corresponds to the multiplication with the complex derivative $\frac{\partial y}{\partial x}$ and $\mathrm{D}F^T$ to the multiplication with its complex conjugate.
Similarly, in the multi-variate case $F:\mathbb{C}^N\to\mathbb{C}^M$, the derivative $\mathrm{D}F:\mathbb{C}^N\to\mathbb{C}^M$ is linear with respect to $\mathbb{C}$ iff $F$ is holomorphic.
If $\mathrm{D}F:\mathbb{C}^N\to\mathbb{C}^M$ is not linear with respect to $\mathbb{C}$, we still call the transposed of the real-valued derivative $\mathrm{D}F^T:\mathbb{R}^{2M}\to\mathbb{R}^{2N}$ the adjoint derivative $\mathrm{D}F^H:\mathbb{C}^M\to\mathbb{C}^N$.

The loss of a neural network in \eqref{eq:network_optimize} must be real as there is no ordering on $\mathbb{C}$.
In the picture of real-valued derivatives, training a network with complex-valued weights is equivalent to optimize the real and imaginary part of the weights independently. In the picture of Wirtinger calculus, we consider a mapping $F:\mathbb{C}^N\to\mathbb{R}$. Since the output is real, it holds $\bar{\frac{\partial F}{\partial x_j}}=\frac{\partial F}{\partial \bar{x_j}}$ such that 
\begin{align}
    \mathrm{D}F^H(1)
    &=
    2\begin{pmatrix}\frac{\partial F}{\partial \bar{x}_1} &\dots& \frac{\partial F}{\partial \bar{x}_N}\end{pmatrix}^T \notag\\
    &=
    \begin{pmatrix}\frac{\partial F_r}{\partial x_{1r}}+i\frac{\partial F_r}{\partial x_{1i}} &\dots& \frac{\partial F_r}{\partial x_{Nr}}+i\frac{\partial F_r}{\partial x_{Ni}}\end{pmatrix}^T\,.
\end{align}
We stress the analogy to \eqref{eq:gradient_transposed}, i.e. the real part of $\mathrm{D}F^H(1)$ is the gradient of $F$ with respect to the real part of $\mathbf{x}$ and the imaginary part of $\mathrm{D}F^H(1)$ is the gradient of $F$ with respect to the imaginary part of $\mathbf{x}$.     \paragraph{\color{black}Implementation of Operators}  For interested programmers, we describe the C-implementation of \texttt{operator}s, \texttt{linop}s and \texttt{nlop}s in the non-linear operator framework.   \texttt{operator}s are the basic structures of the framework.
An \texttt{operator} holds an apply-function which is called when the operator is applied and a generic data structure which is passed to this function together with pointers to the input and output md-arrays (\figref[A]{fig:operatorstructures}). 
An example for an \texttt{operator} is the chain-\texttt{operator} (\figref[D]{fig:operatorstructures}, whose data structure holds references to the   chained \texttt{operator}s and whose apply function calls them one after another.  A \texttt{linop} $A:\mathbb{C}^N\to \mathbb{C}^M$ models a linear operator by holding references to \texttt{operator}s  computing $A$ and its adjoint $A^H$.  For atomic, i.e. non-composed, \texttt{linop}s, the \texttt{operator}s have  access to a shared data structure of type \texttt{linop\_data\_s} (\figref[B]{fig:operatorstructures}).      For example, a \texttt{linop} performing a matrix multiplication stores the matrix in this structure such that it can be accessed by the forward and adjoint \texttt{operator}s.  \texttt{linop}s are chained by creating a new \texttt{linop} referring to the chained forward and adjoint \texttt{operator}s.
An \texttt{nlop} consists of an \texttt{operator} modeling the non-linear forward operator and \texttt{linop}s modeling the derivatives.
For atomic \texttt{nlop}s, the \texttt{linop}s and the forward \texttt{operator}, have access to a shared data structure of type \texttt{nlop\_data\_s} (c.f. \figref[C]{fig:operatorstructures}) to store the data necessary to evaluate the derivatives at the last input of the forward operator as described above.   To implement a completely new \texttt{nlop}, the programmer needs to define the data structure \texttt{nlop\_data\_s} and functions to be called when the \texttt{nlop} or its (adjoint) derivative is applied.
Other references and data structures are created automatically.  All shared data structures use reference counting for automatic memory management (garbage collection).  As \texttt{nlop}s are implemented based on md-functions, they are automatically executed
on the GPU if the inputs and outputs are located on the GPU.   %The basic structure of atomic, i.e. non-composed, \texttt{operator} is depicted in \figref[A]{fig:operatorstructures}.
%An \texttt{operator} holds a reference to a data structure \texttt{data} and a function \texttt{apply} which is called when the operator is applied.
%The data structure can for example hold a matrix, if the operator is used to multiply an input vector with a matrix.
%When the \texttt{operator} is applied, the apply function is passed references to pre-allocated memory for the inputs and outputs and the data structure of the operator.

%A \texttt{linop} $A:\mathbb{C}^N\to \mathbb{C}^M$ models a linear operator.
%\texttt{linop}s can be used to apply the linear operator $A$ and its adjoint $A^H$, where both, $A$ and $A^H$, are implemented as independent \texttt{operator}s.
%Moreover, a \texttt{linop} can contain an \texttt{operator} for an optimized implementation of the chained forward and adjoint operator $A^HA$ or its regularized inverse.
%The four \texttt{operator}s of an atomic \texttt{linop} have access to a shared data structure of type \cd{linop_data_s}. (c.f. \figref[B]{fig:operatorstructures}).
%Thus, data which is needed to apply the forward or the adjoint of a \texttt{linop}, for example a matrix, needs to be stored only once.

\begin{figure}
    \centering
    \includegraphics[width=.7\linewidth]{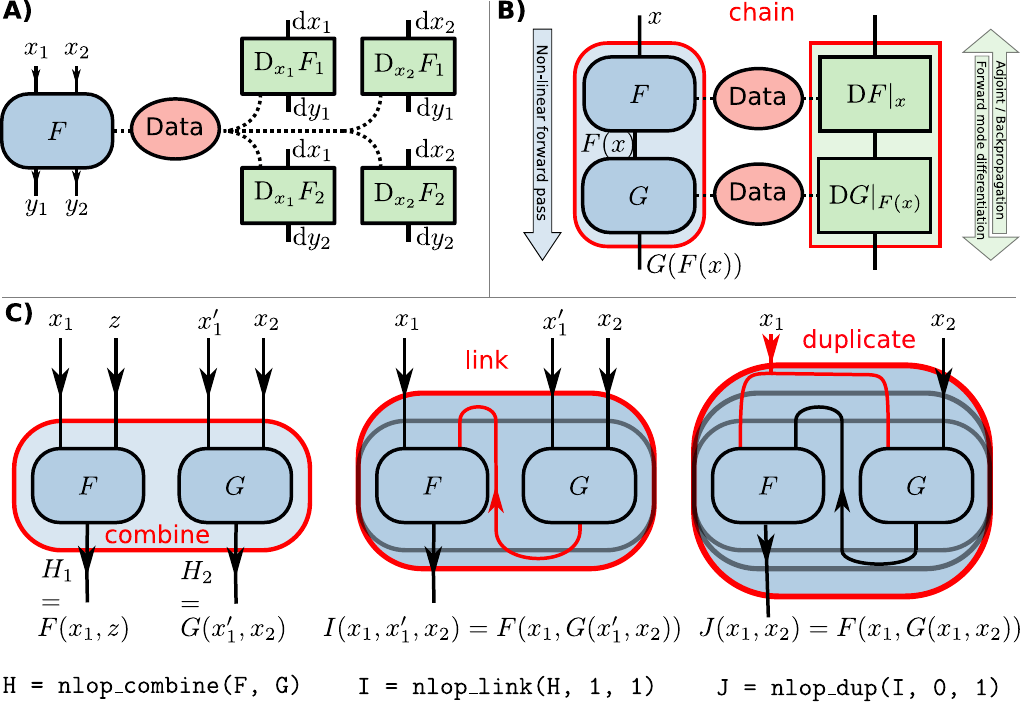}
    \caption{
        Basic concepts of \texttt{nlop}s.
        \textbf{A)} An atomic \texttt{nlop} exemplary with two complex-valued inputs ($\mathbf{x}_1$, $\mathbf{x}_2$) and two outputs ($\mathbf{y}_1=F_1(\mathbf{x}_1,\mathbf{x}_2)$, $\mathbf{y}_2=F_2(\mathbf{x}_1,\mathbf{x}_2)$) consisting of the forward \texttt{operator} $F$ and its derivatives $\mathrm{D}_i{F_o}$ modeled by \texttt{linop}s.          $F$ and $\mathrm{D}_i{F_o}$ communicate via a shared data structure.         \textbf{B)} Chaining of two \texttt{nlop}s $F$ and $G$.                  Since $G$ is applied on the output $F(\mathbf{x})$, its derivative $\mathrm{D}G|_{F(\mathbf{x})}$ is automatically evaluated at $F(\mathbf{x})$.                   \textbf{C)} The two \texttt{nlop}s $F$ and $G$ are combined to form $H$, whose output 1 is linked into input 1 to form $I$, whose inputs 0 and 1 are duplicated to construct $J(\mathbf{x}_1,\mathbf{x}_2)=F(\mathbf{x}_1, G(\mathbf{x}_1, \mathbf{x}_2))$.         The derivatives of the final operator are constructed automatically (not shown).
    }
    \label{fig:nlops}
\end{figure}

\begin{figure}
    \centering
    \includegraphics[width=.7\linewidth]{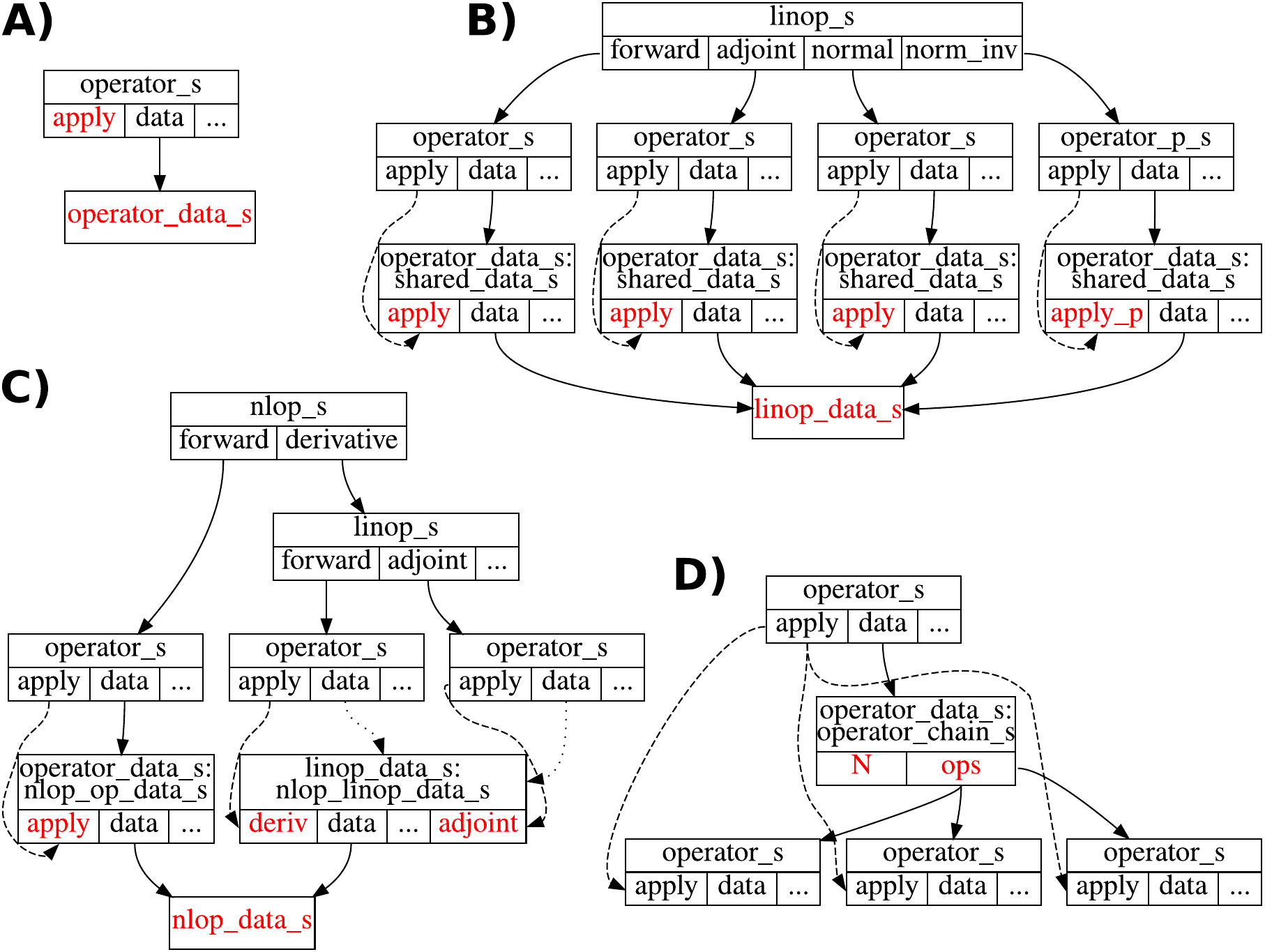}
    \caption{
        Schematic description of \texttt{operator}s, \texttt{linop}s and \texttt{nlop}s as data structures in BART.
        Solid lines mean ``points to'', dotted lines ``points to indirectly'' and dashed lines ``calls''. Colons indicate specific realizations of a data structure, i.e. \texttt{operator\_chain\_s} is the \texttt{operator\_data\_s} structure used for chaining operators.         Objects required to create the respective structures are marked in red. Other structures and references are created automatically.         \textbf{A)} An \texttt{operator} holds a reference to a data structure and a function which is called when the \texttt{operator} is applied.
        \textbf{B)} A \texttt{linop} holds references to multiple \texttt{operator}s such as the forward and adjoint operator which share a common data structure.
        \textbf{C)} An \texttt{nlop} holds references to the non-linear forward \texttt{operator} and \texttt{linop}s modeling the derivatives.
        The forward \texttt{operator} and \texttt{linop}s have access to a shared data structure \texttt{nlop\_data\_s}.
        \textbf{D)} The data structure of a chain-\texttt{operator} holds references to the chained \texttt{operator}s which are applied sequentially, when the chain-\texttt{operator} is applied.
}
    \label{fig:operatorstructures}
\end{figure}

    \paragraph{Functional Container}

Generally, the execution properties of an \texttt{nlop} can be modified by encapsulating it in a container which itself is an \texttt{nlop}.
We use such a container to implement checkpointing to reduce memory use.
When the checkpointing-container is applied, the inputs are stored and the inner \texttt{nlop} is applied without saving data for computing its derivatives.
When the derivatives are needed, the inner \texttt{nlop} is applied again using the inputs stored in the container and the data needed for the derivatives is re-computed.
Thus, checkpointing can reduce the memory consumption at the price of multiple applications of the \texttt{nlop}.

Moreover, a functional container can be used to assign an \texttt{nlop} to a specific GPU.
When such an \text{nlop} or its derivative is called, the CUDA context is switched to the selected GPU and all input data of the \texttt{nlop} are copied to the GPU.
Afterwards the inner \texttt{nlop} is called which uses the selected GPU for all its computation and memory allocations.
By calling \texttt{nlop}s assigned to different GPUs from different threads in parallel, we can efficiently distribute the memory
and computation of the \texttt{nlop}s to multiple GPUs (c.f. Supporting Information \figref{supfig:multi_gpu}).

\subsubsection{Neural Network Library}

%For convenient construction of neural networks based on \texttt{nlop}s, we implemented the neural network (nn)-library.
%The nn-library allows to index the arguments (inputs and outputs) of \texttt{nlop}s by meaningful names instead of numeric indices.
%The arguments are annotated by a type defining how the optimization algorithm treats this argument (weights, data, moving statistics of batch normalization) and inputs corresponding to weights can be attached with an initializer.

The neural network (nn)-library contains our complex-valued implementations of typical operators used to construct neural networks, i.e.
\begin{itemize}
	\item fully-connected (dense) layers
	\item (transposed / adjoint) convolutional layers
	\item dropout layers, max-pooling layers, batch normalization layer \cite{Ioffe__2015}
	\item activation layers: complex cardioid\cite{Virtue__2017}, $\mathbb{C}$ReLU\cite{Trabelsi__2018, Cole_Magn.Reson.Med._2021}, sigmoid and softmax
	\item loss functions: mean squared error (MSE), mean absolute difference, structural similarity index measure (SSIM)\cite{Zhao_IEEETrans.Comput.Imag._2017}, generalized dice loss \cite{Sudre__2017} and categorical cross-entropy.
\end{itemize}  The corresponding \texttt{nlop}s are implemented generically such that they act on N-dimensional complex-valued md-arrays and support operations (convolution, normalization, pooling) along arbitrary dimensions.
However, currently convolutions are only backed by optimized GPU code for up to 3 dimensions.      Moreover, the nn-library contains another wrapper for \texttt{nlop}s to index the arguments (inputs and outputs) of \texttt{nlop}s by meaningful names instead of numeric indices.
The arguments are annotated by a type defining how the optimization algorithm treats this argument (weights, data, moving statistics of batch normalization) and inputs corresponding to weights can be attached with an initializer and proximal operators for regularization.   \paragraph{Integration of TensorFlow Graphs}

The \texttt{nlop} framework also serves as a generic wrapper for computation graphs exported from other deep learning frameworks.
As a proof of concept, we have implemented a wrapper for TensorFlow graphs based on the TensorFlow C API\footnote{\url{https://www.tensorflow.org/install/lang_c}}.
A pre-trained neural network based on TensorFlow can be exported to a graph file which is imported by BART to construct an \texttt{nlop}.
When this \texttt{nlop} is applied, the forward-pass of the TensorFlow graph is executed, while TensorFlow's gradients are used to compute the adjoint derivative of the \texttt{nlop}. 
%The adjoint derivative of the \texttt{nlop} corresponds to the backward-pass of the TensorFlow graph, i.e. the computation of the gradients.
%In principle, the TensorFlow graph could be annotated with the required gradients automatically using the TensorFlow C API, however, since C API does not support all gradients, the user is required to annotate the gradients manually before the graph is exported.
The forward derivative for the TensorFlow wrapper is not implemented.

\subsubsection{Iterative Training Algorithms}

Training a neural network corresponds to minimizing the loss $\sum_i L(\mathbf{y}_i, F(\mathbf{x}_i;\mathbf{\theta}))$ with respect to the weights $\mathbf{\theta}$ (c.f. \eqref{eq:network_optimize}).
Having constructed an \texttt{nlop} representing $F$, we chain its output into another \texttt{nlop} $L$ to generate a loss-\texttt{nlop}.
This loss-\texttt{nlop} has two types of inputs, those corresponding to weights $\mathbf{\theta}$ and those corresponding to data $\mathbf{x}$, $\mathbf{y}$.
The training dataset $\mathbf{x}_i,\mathbf{y}_i$ is split into mini-batches and in each iteration the weights $\mathbf{\theta}$ are updated based on the gradient with respect to these weights.
For training neural networks, we have integrated incremental gradient methods such as stochastic gradient descent, Adam \cite{1412.6980}, and iPALM \cite{Pock_SIAMJ.Img.Sci._2016} into BART's library for iterative algorithms.
%These algorithms are accessed using a unified interface provided, amongst other things, the loss-\texttt{nlop}, flags describing the type of the respective inputs (i.e. if an input corresponds to data or weights), the training data and the pre-initialized memory for the weights.

\subsection{Applications and Implemented Networks}

To demonstrate practicability of our framework, we have implemented and trained VarNet \cite{Hammernik_Magn.Reson.Med._2017} and MoDL \cite{Aggarwal_IEEETrans.Med.Imag._2019}.
Both networks are motivated by unrolling an optimization algorithm solving the inverse problem
\begin{equation}
    \mathbf{x}^*=\argmin_{\mathbf{x}} \norm{A\mathbf{x}-\mathbf{y}}^2 + R(\mathbf{x})\,.
    \label{eq:inverse_problem}
\end{equation}
Here, $A=\mathcal{PFC}$ is the linear SENSE operator composed of the multiplication with the $\mathcal{C}$oil sensitivity maps, the $\mathcal{F}$ourier transform, and the projection to the sampling $\mathcal{P}$attern.
$\mathbf{x}$ is the MR image to be reconstructed and $\mathbf{y}$ is the measured k-space data.
$R(\mathbf{x})$ is a regularization term imposing prior knowledge on the reconstructed image $\mathbf{x}$.

We first describe the structure of both networks and our respective implementations.
Afterwards, we describe how the TensorFlow wrapper can be used to integrate an externally trained regularizer $R(\mathbf{x})$ for reconstruction with BART.
Scripts to reproduce training and application of the networks are available at \url{https://github.com/mrirecon/deep-deep-learning-with-bart}.  To provide interested developers a starting point to implement neural networks in BART, we have implemented a toy network to classify handwritten digits of the MNIST\cite{mnist_database} database.
The network can be found in the BART source code at \texttt{src/mnist.c} and scripts to prepare the MNIST database are available in the script repository.   \subsubsection{Variational Network}

VarNet is motivated by solving \eqref{eq:inverse_problem} using an unrolled gradient descent algorithm that includes a trained regularizer $R$.
The network is initialized with the adjoint reconstruction $\mathbf{x}^0=A^H\mathbf{y}$ and updates the reconstruction $\mathbf{x}^t$ by 
\begin{align}
    \mathbf{x}^{t+1}=\mathbf{x}^t-\sum_{i=1}^{N_k}(K_i^t)^T\Phi^{t'}_i\left(K_i^t\mathbf{x}^t\right) - \lambda^t\left(A^HA\mathbf{x}^t-A^H\mathbf{y}\right)\notag\\ 0 \le t\le T-1\,.\label{eq:update_vn}
\end{align}
Here, the sum corresponds to the gradient of a regularizer $R^t(\mathbf{x})=\sum_{i=1}^{N_k}\Phi^t_i(K_i^t\mathbf{x})$, where $K$ is a convolution with $N_k$ filters and $\Phi'$ is the derivative of a trainable activation function.
The imaginary part of the convolved images $K_i^t\mathbf{x}$ is discarded to be consistent with the original implementation of VarNet.
The last term corresponds to a gradient step of the data-consistency term $\norm{A\mathbf{x}-\mathbf{y}}^2_2$ with trained step size $\lambda^t$.   The BART implementation of VarNet can be trained and applied with the \texttt{reconet} command of the BART toolbox, i.e.
\begin{lstlisting}
$ bart reconet --network=varnet --train <kspace> <coils> <weights> <reference>
$ bart reconet --network=varnet --apply <kspace> <coils> <weights> <output>
\end{lstlisting}      \texttt{$<$kspace$>$}, \texttt{$<$coils$>$} and \texttt{$<$reference$>$} are input files holding multi-dimensional arrays as training data or for inference.
The data layout follows the BART convention and stacks independent datasets/volumes along the batch dimension 15.
    An undersampling pattern can be provided to subsample the k-space, otherwise the pattern is estimated from the k-space data.       \texttt{$<$weights$>$} is a file holding the network weights $\mathbf{\theta}$ which are an output of the training command and an input for the reconstruction/apply command.      Further options, such as network parameters, training losses , initializations for the weights, or the training algorithm can be configured using command line options.
The default hyperparameter are based on the TensorFlow implementation\footnote{\url{https://github.com/VLOGroup/mri-variationalnetwork}, Commit: 4b6855f}, i.e. $T=10$ iterations, $N_k=24$ $11\times11$-convolution filter and $N_w=31$ Gaussian radial basis functions to construct the activation $\Phi'$. This results in 65,530 real-valued trainable parameters. iPALM is used as training algorithm.
The \texttt{--normalize} option can be used to scale the data such that $1=\max{\left|\mathbf{x}^0\right|}$.
As our implementation is equivalent to the original one, weights trained with TensorFlow can be exported for inference with BART.

\subsubsection{MoDL}
The MoDL \cite{Aggarwal_IEEETrans.Med.Imag._2019} network is another unrolled network initialized with $\mathbf{x}^0=A^H\mathbf{y}$.
A residual network $\mathcal{D_W}$ denoises the current reconstruction and data-consistency is imposed by a proximal mapping.
The iterations read
\begin{align}
    \mathbf{x}^{t+1} &= \argmin_{\mathbf{x}}\left[\norm{A\mathbf{x}-\mathbf{y}}^2+\lambda\norm{\mathbf{x}-\mathcal{D_W}(\mathbf{x}^{t})}^2\right]\notag\\
    &= \underbrace{\left(A^HA+\lambda\id \right)^{-1}}_{\mathcal{Q}}\left(A^H\mathbf{y}+\lambda \mathcal{D_W}(\mathbf{x}^{t})\right)\quad0\le t\le T-1\,.\label{eq:update_modl}
\end{align}
The residual network $\mathcal{D_W}$ consists of $L$ convolutional layers with $F$ filters followed by batch normalization layers and ReLU activation functions.
The \texttt{reconet} command with the \texttt{--network=modl} option is used to train MoDL.
By default, the Adam algorithm and the hyperparameter from the TensorFlow implementation\footnote{\url{https://github.com/hkaggarwal/modl}, Commit: 428ef84} are used, i.e. the network is unrolled for $T=10$ unrolled iterations with shared weights and each residual block contains $L=5$ convolutional layers with $F_{\mathrm{c}}=32$ complex-valued filters (instead of $F_{\mathrm{r}}=64$ real-valued in the TensorFlow implementation). Thus, our implementation has 28,364 complex-valued (= 56,728 real-valued) trainable parameters in contrast to 112,001 real-valued parameters in the Tensor-Flow implementation.  To compute $\mathcal{Q}(\mathbf{x})$, we have implemented a generic inversion module for positive-definite self-adjoint linear operators $S_\lambda:\mathbb{C}^N\to\mathbb{C}^N$ parametrized with parameters $\lambda \in \mathbb{C}^M$.
%In case of MoDL, $S_\lambda=A^HA + \lambda\id$ and $\lambda$ is just one real-valued parameter, however, the same implementation could be for example used to interpret the coil sensitivity maps as parameters of $A$ as needed for networks jointly estimating coil sensitivity maps and image content \cite{Arvinte_MICCAI_2021}.
%This would allow the computation of the derivatives of $\mathcal{Q}\mathbf{x}$ with respect to the coil sensitivity maps as needed for networks jointly estimating the coil sensitivity maps and image content \cite{Arvinte_MICCAI_2021}.
Given an \texttt{nlop} $\mathcal{S}$ that has $\lambda$ as second input and applies the parametrized linear operator $S_\lambda$ to its first input,
we construct the \texttt{nlop} $\mathcal{S}^{-1}$ applying the inverse $S^{-1}_\lambda$.
These \texttt{nlops} are defined by
\begin{align}
	\mathcal{S}: &\mathbb{C}^N \times \mathbb{C}^M \to \mathbb{C}^N & \mathcal{S}^{-1}: &\mathbb{C}^N \times \mathbb{C}^M \to \mathbb{C}^N \notag\\
	&(\mathbf{x}, \lambda)\mapsto \mathbf{y}=S_\lambda\mathbf{x} && (\mathbf{y}, \lambda)\mapsto \mathbf{x}=S_\lambda^{-1}\mathbf{y}\;.
\end{align}
Note that the derivatives of $\mathcal{S}^{-1}$ can be expressed in terms of $\mathcal{S}$ and its derivatives, i.e.
\begin{align}
	\mathrm{D}_{\mathbf{y}}\mathcal{S}^{-1}|_{\mathbf{y}, \lambda}:&\mathrm{d}\mathbf{y} \mapsto \mathrm{d}\mathbf{x} =  S_\lambda^{-1} \mathrm{d}\mathbf{y} \notag\\
	\mathrm{D}_{\lambda}\mathcal{S}^{-1}|_{\mathbf{y}, \lambda}:&\mathrm{d}\lambda \mapsto \mathrm{d}\mathbf{x} = -S_\lambda^{-1} \circ \mathrm{D}_\lambda\mathcal{S}|_{\mathcal{S}^{-1}(\mathbf{y}, \lambda), \lambda} \mathrm{d}\lambda\;.
\end{align}
As proposed by Aggarwal et al. \cite{Aggarwal_IEEETrans.Med.Imag._2019}, we use the conjugate gradient algorithm to apply $S_\lambda^{-1}$.
%The derivative with respect to the parameters $\lambda$ involves the derivative of $\mathcal{S}$ evaluated at the result of the forward path $\mathbf{y} = \mathcal{S}^{-1}(\mathbf{x}, \lambda)$.
%This derivative is computed using the \texttt{nlop} $\mathcal{S}$.

\subsubsection{Extensions to the SENSE-Model}

BART's implementation of the SENSE model is generic in the sense that it can handle multiple sets of coil sensitivity maps (soft-SENSE \cite{Uecker_Magn.Reson.Med._2014})
and supports non-Cartesian sampling patterns.
The soft-SENSE model is suitable if the object exceeds the FOV since one set of coil sensitivity maps can not explain infolding artifacts \cite{Uecker_Magn.Reson.Med._2014}.
In the context of deep-learning, the soft-SENSE model has been used recently\cite{Sandino_Magn.Reson.Med._2020, Hammernik_Magn.Reson.Med._2021,Johnson_J.Magn.Reson.Imaging_2021}.
VarNet and MoDL only update the image corresponding to the first set of coil sensitivity maps in the network block.
We use the MSE loss on the coil images since the coil images serve as a reference independent
of the estimated coil sensitivity maps.
%FIXME: Do we need an equation?

Reconstruction networks for non-Cartesian sampling trajectories have been investigated recently \cite{Sandino_Magn.Reson.Med._2020,Schlemper__2019, Kofler_Med.Phys._2021, Ramzi__2021}. % where \cite{Kofler_Med.Phys._2021,Ramzi__2021} make use of the PyTorch implementation of the non-uniform FFT \cite{Muckley__2020}.
BART implements the non-uniform (nu)FFT as a \texttt{linop} which is integrated in the SENSE model of VarNet and MoDL.
%The nuFFT in Eqs.~\ref{eq:update_vn} and \ref{eq:update_modl} is needed once to compute the adjoint $A^H\mathbf{y}$ and afterwards only in the combined forward and adjoint operator $\mathcal{F}^H\mathcal{P}^H\mathcal{P}\mathcal{F}$.
To save expensive gridding steps of the nuFFT, we precompute the adjoint reconstruction $A^H\mathbf{y}$ and the point spread function (PSF) of the non-Cartesian sampling pattern $\mathcal{P}$ for the whole dataset.
The joint forward-backward nuFFT $\mathcal{F}^H\mathcal{P}^H\mathcal{P}\mathcal{F}$ is implemented by the convolution with the PSF (Toeplitz trick) \cite{Wajer__2001,Fessler_IEEETrans.SignalProcessing_2005}, which significantly speeds up computations on the GPU \cite{Uecker_Magn.Reson.Med._2010, Baron_Magn.Reson.Med._2017}.
We initialize the non-Cartesian networks with a SENSE reconstruction $\mathbf{x}^0=\mathcal{Q} A^H \mathbf{y}$.
For MoDL, the number of CG-iterations in each data-consistency block has been increased from 10 to 30, while the number of unrolled iterations has been reduced to $T=5$.

\subsubsection{Image Reconstruction using a Learned Prior}

An alternative approach of using neural networks for MRI reconstruction is learning prior knowledge about the image distribution by learning a regularizer $R(\mathbf{x})$ independently of the reconstruction.
For reconstruction, the learned regularizer is inserted into \eqref{eq:inverse_problem}.
One approach to learn a regularizer is based on deep Bayesian estimation
\cite{Luo_Magn.Reson.Med._2020}. The resulting regularizer is given by
\begin{equation}
    R(\mathbf{x}) = -\lambda \log p(\mathbf{x};\mathtt{Net}(\mathbf{x}, \mathbf{\theta}^*))\,.
    \label{eq:reg_pixelcnn}
\end{equation}
Here, $\mathtt{Net}(\mathbf{x}, \mathbf{\theta}^*)$ denotes the PixelCNN++ \cite{Salimans_arXiv_2017} which is trained to predict the conditioned distribution parameters of the mixture of logistic distributions that are used to model the image distribution.
Inserting the regularizer $R$ defined in \eqref{eq:reg_pixelcnn} into the optimization problem \eqref{eq:inverse_problem} corresponds to a maximum a posteriori estimation for the reconstructed image.
For more details, we refer to the original publication \cite{Luo_Magn.Reson.Med._2020}.

For reconstruction, the trained TensorFlow graph computing $R(\mathbf{x})$ is exported and loaded into BART using the TensorFlow wrapper described above.
The resulting \texttt{nlop} is used to construct the corresponding proximal operator
\begin{equation}
    \mathrm{prox}_R(\mathbf{v}) = \argmin_{\mathbf{x}} \frac{1}{2} \norm{x-v}^2 + R(\mathbf{x})\,.
\end{equation}  The proximal operator is computed by the gradient descent algorithm using automatic differentiation to compute $\nabla R$.  The proximal operator can be plugged into any of BART's proximal operator based iterative optimization algorithms using the \texttt{pics} command, i.e. 
\begin{lstlisting}
$ bart pics -R TF:{model_path}:lambda <kspace> <coils> <output>
\end{lstlisting}

\section{Results}
\label{sed:results}

\subsection{Reconstructions with BART}

\begin{figure*}
	\includegraphics[width=\linewidth]{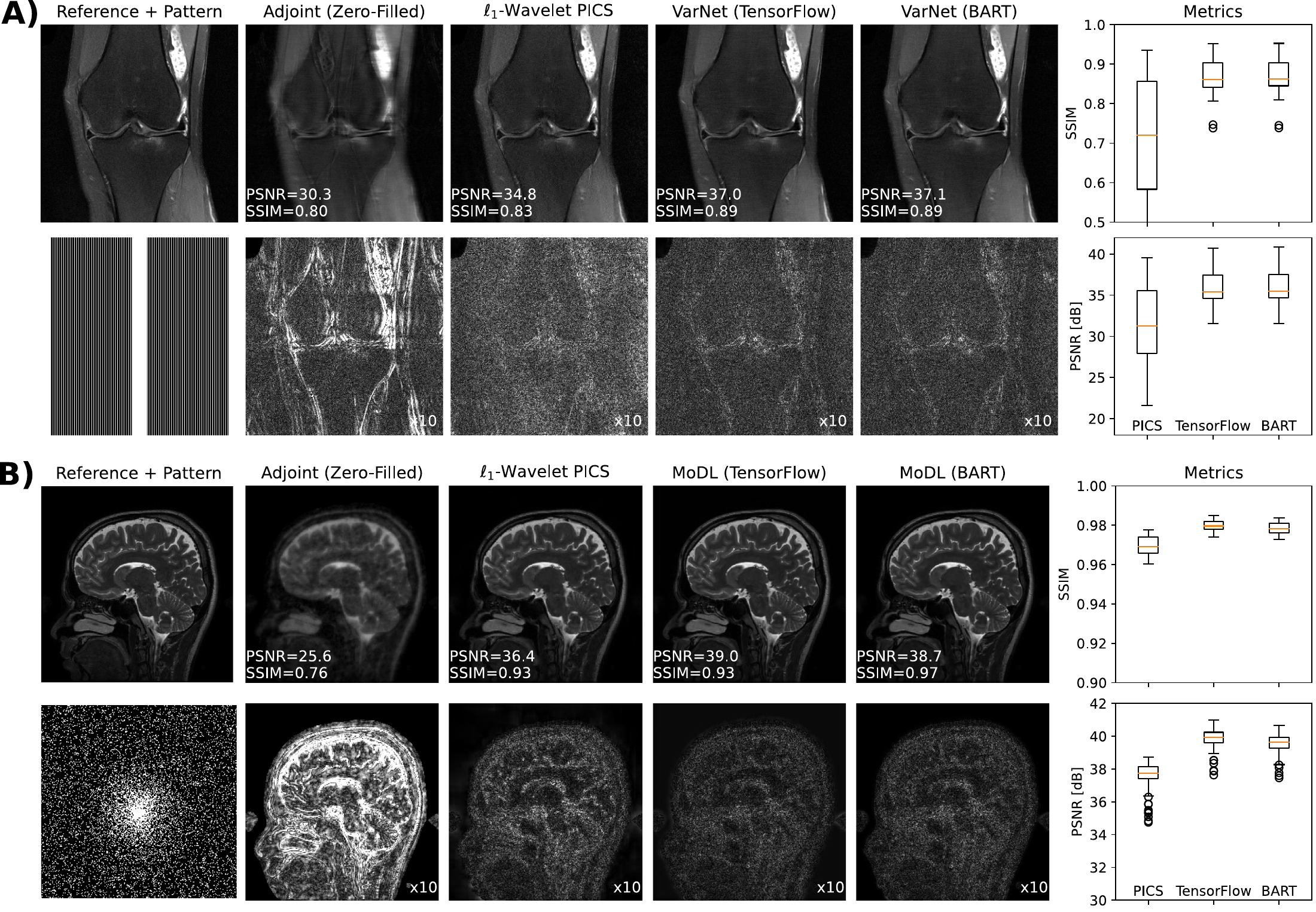}
	\caption{
        Comparison of the TensorFlow and BART implementation of VarNet (A) and MoDL (B). For reference, we also show the results of the adjoint reconstruction $A^H\mathbf{y}$ and an $\ell_1$-Wavelet regularized SENSE reconstruction computed using the BART pics tool.                           Boxplots are based on PSNR and SSIM of the respective evaluation datasets using the coil sensitivities as foreground mask.
        This mask explains the discrepancy to the SSIM values given at the reconstructed images.                   }
	\label{fig:vn_modl_cmp}
\end{figure*}

%In this section, we present example reconstructions using our trained networks and compare them with reconstructions based on the original TensorFlow implementations.
We have trained VarNet and MoDL on the datasets provided with the respective publications using the BART and TensorFlow implementations to compare the reconstruction quality.
%In the following, the dataset published with the VarNet is referred to as knee-dataset and the dataset published with MoDL is referred to as brain-dataset.
VarNet was trained for 30 epochs algorithm with a batch size $N_\mathrm{b}=10$ on 300 randomly ordered slices of 15 subjects from the \texttt{coronal\_pd\_fs} directory, while 20 slices of the remaining 5 subjects were used for evaluation.  The 15-coil fully-sampled k-space data was retrospectively subsampled (4-fold regular undersampling, 28 auto calibration lines).  MoDL was trained with batch size $N_\mathrm{b}=10$ in a two-step-approach, i.e. the weights were initialized for 100 epochs using $T=1$ unrolled iteration and afterwards the network was trained for 50 epochs with $T=10$.  The brain dataset of MoDL consists of 5 subjects acquired with a 12 channel head coil.
90 slices of the first 4 subjects (360 in total) were used for training and 100 slices of the remaining subject for testing.
Subsampled k-space data was generated from the fully-sampled images by multiplying them with provided coil sensitivity maps, Fourier transform, subsampling (variable density with acceleration 6, no auto-calibration region) and addition of Gaussian noise with standard deviation $\sigma = 0.001$.
This procedure is used in the TensorFlow implementation to produce training data.     We only used normalization as described above for the knee data of VarNet.  %The default network parameters described in the methods part were used for both networks and only the input data of VarNet were normalized.
We show example reconstructions based on the respective networks and implementations in \figref{fig:vn_modl_cmp}.
Both implementations of the respective networks perform quite similar and better than the classical $\ell_1$-Wavelet regularized reconstruction.
%The TensorFlow implementation has a lower SSIM in the example reconstruction due to an offset in the background.
%To have a fair comparison based on the image content, we used the coil sensitivity maps as a foreground mask to compute the PSNR and SSIM of the respective evaluation datasets.
To support this statement quantitatively, we computed the PSNR and SSIM for each slice in the evaluation dataset and visualize the results in the boxplots in \figref{fig:vn_modl_cmp}.
Moreover, we compare in Supporting Information \figref{supfig:train_history} the mean PSNR and SSIM of the VarNet evaluation dataset computed after each training epoch.
%Both implementations of the respective networks perform similar in terms of PSNR and SSIM.

\begin{figure*}
    \centering
	\includegraphics[width=\linewidth]{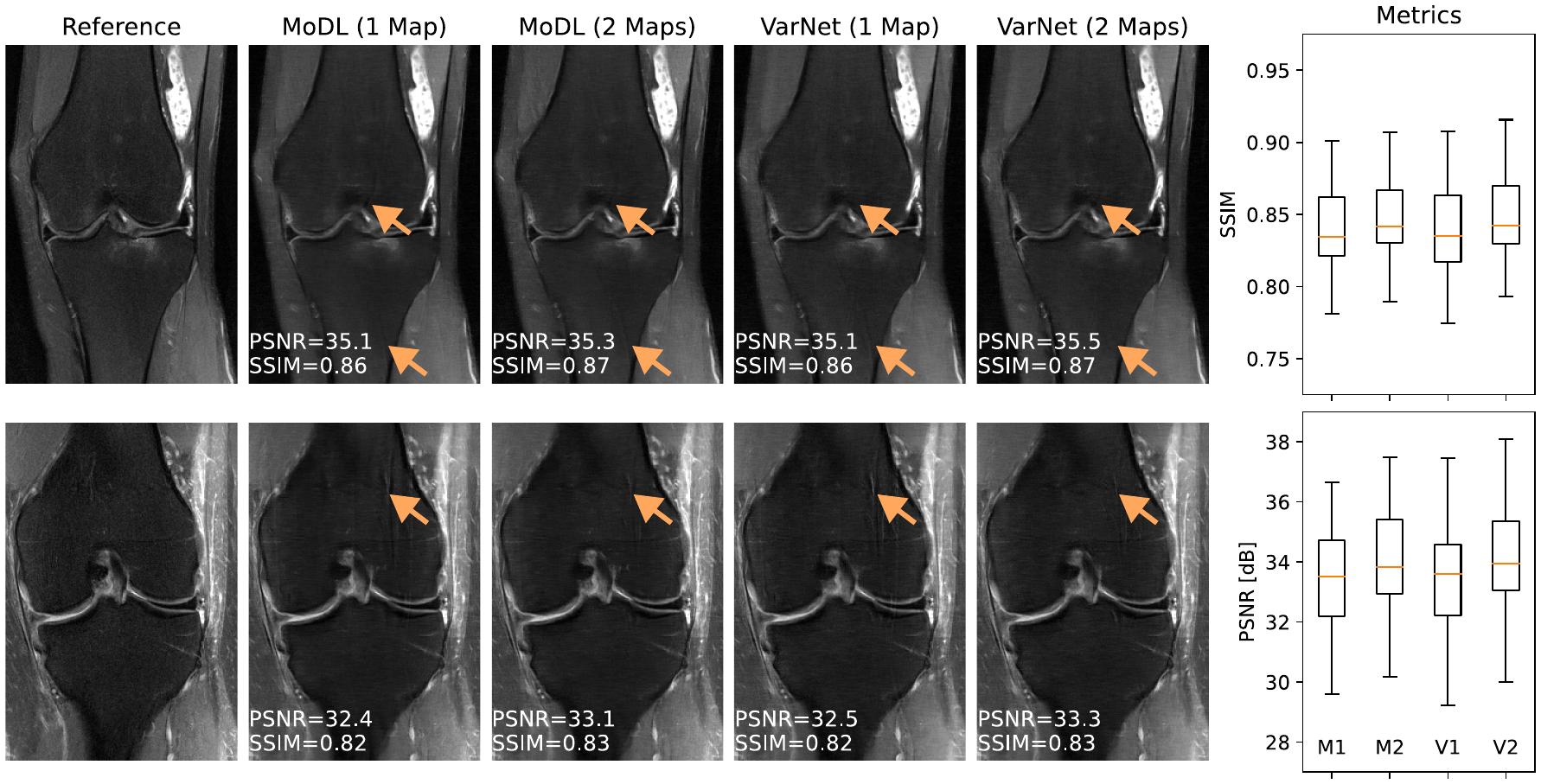}
	\caption{
        Comparison of two example reconstructions with MoDL and VarNet using one set of coil sensitivity maps (usual SENSE) and two sets of coil sensitivity maps (soft-SENSE).
        The aliased k-space data is simulated by first zero-padding the fully-sampled coil-images and afterwards sub-sampling the k-space by a factor of two before applying the usual sampling pattern (every fourth line and 28 auto calibration lines).
        The usage of two sets of coil sensitivity maps reduce undersampling artifacts (c.f. arrows) and improves the PSNR and SSIM for VarNet and MoDL.
        %For the first example, one set of coil sensitivity maps should be sufficient as the knee fits into the field of view.
        %We speculate that the two-maps reconstruction outperforms the one-map reconstruction due to a more efficient training.
    }
	\label{fig:softsense}
\end{figure*}

To demonstrate the benefits of the soft-SENSE model in the case that the object exceeds the FOV, we simulated k-space data with a reduced FOV from the fully-sampled knee dataset and trained VarNet and MoDL on this dataset.
ESPIRiT \cite{Uecker_Magn.Reson.Med._2014} was used to estimate either one or two sets of coil sensitivity maps from the simulated k-space.
Respective example reconstructions and quantitative metrics computed on the evaluation dataset are presented in \figref{fig:softsense}.
Reconstructions using two sets of coil sensitivity maps show less aliasing artifacts and are superior in terms of SSIM and PSNR.

\begin{figure*}
    \centering
    \includegraphics[width=\linewidth]{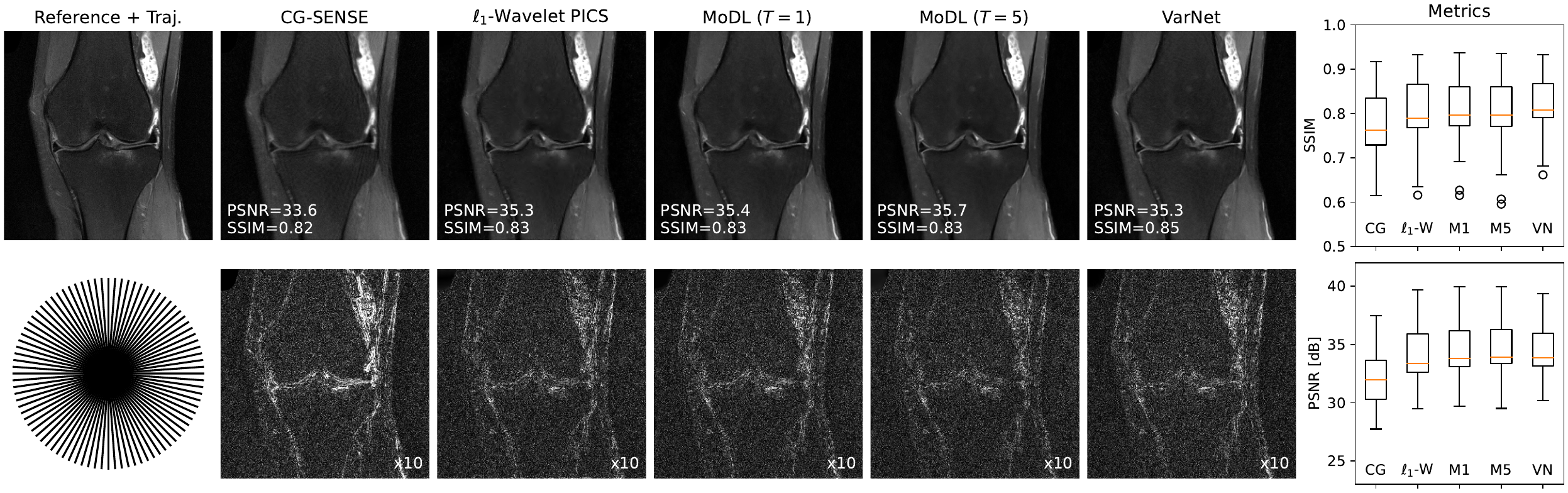}
    \caption{Comparison of MoDL and VarNet for non-Cartesian reconstructions using a radial trajectory with 44 spokes. The fully-sampled k-space data from the reference knee image in \figref{fig:vn_modl_cmp} was interpolated on the trajectory to simulate the non-Cartesian k-space data. For reference, we show the results of the adjoint reconstruction $A^H\mathbf{y}$ with density compensation, a CG-SENSE and $\ell_1$-Wavelet regularized reconstruction computed using the BART pics tool.}
    \label{fig:radial}
\end{figure*}

Further, we used the knee dataset to simulate non-Cartesian k-space data (radial trajectory with 44 spokes) and trained the network on this simulated data.
In \figref{fig:radial}, we present an example reconstruction using the non-Cartesian versions of VarNet and MoDL.
The learned methods improve the image quality compared to the classical $\ell_1$-Wavelet regularized reconstruction slightly.

\subsection{Computational Performance}
\begin{figure}
    \centering
    \includegraphics[width=.7\linewidth]{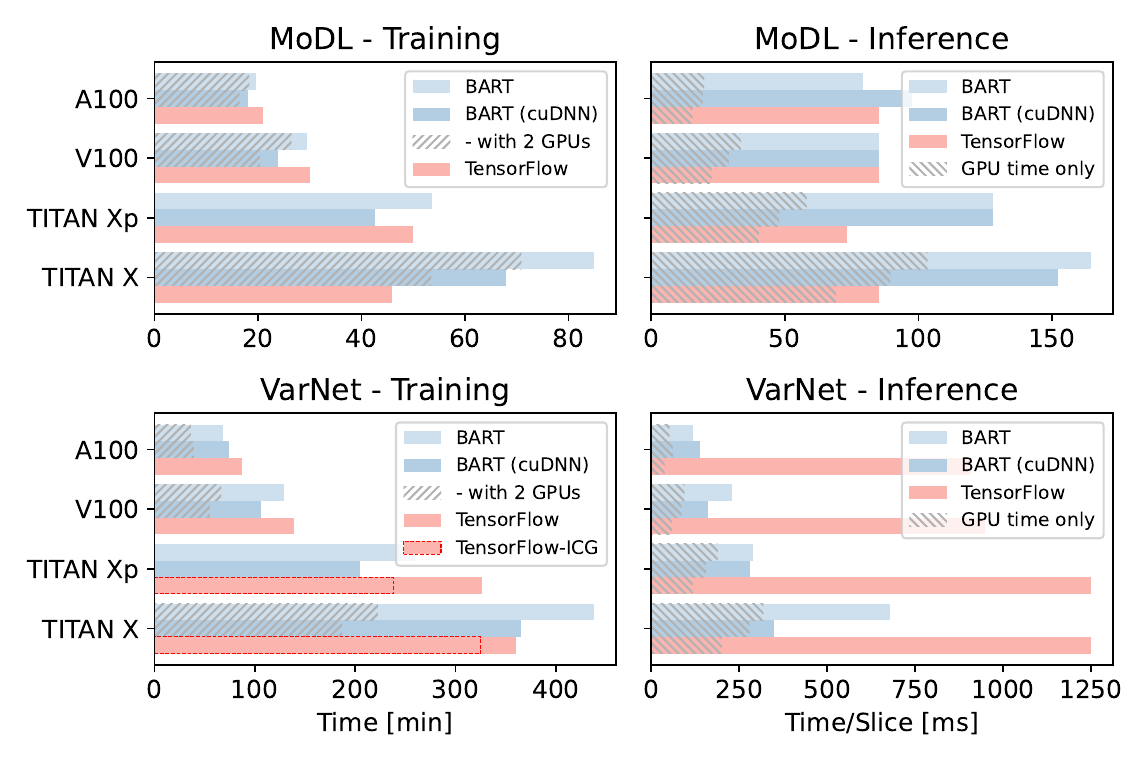}
    \caption{Comparison of training (left) and inference (right) time for MoDL and VarNet on different GPUs (full names in text).          We observed slow host to device copies on the TITAN Xp which might affect the TensorFlow result of MoDL on this GPU.
    In general, the BART and TensorFlow implementations provide similar performance.}
    \label{fig:timing}
\end{figure}

  We compared the BART and the TensorFlow\footnote{VarNet: \url{https://github.com/VLOGroup/mri-variationalnetwork}, Commit: 653630b; MoDL: \url{https://github.com/hkaggarwal/modl}, Commit: 428ef84} implementations of MoDL and VarNet in terms of training time, inference time, and memory consumption on four different Nvidia GPUs, i.e. A100-SXM-80GB, Tesla V100-SXM2-32G, TITAN Xp (12GB) and GTX TITAN X (12GB).  We use TensorFlow 1.15 maintained by NVIDIA to support current GPUs \footnote{https://github.com/NVIDIA/tensorflow} with cuBLAS 11.7, and cuFFT 10.6, cuDNN 8.3.
For VarNet we also experimented with the implementation\footnote{\url{https://github.com/VLOGroup/mri-variationalnetwork}, Commit: 4b6855f} based on TensorFlow-ICG\footnote{https://github.com/VLOGroup/tensorflow-icg; CUDA 8.0; cuDNN 7.0}.
The training time of BART was measured in two settings: once with CUDA 11.2, cuDNN 8.3 and the use of non-deterministic algorithm and once without cuDNN and only using deterministic algorithms.
All network parameters were chosen as described before except the number of unrolled iterations of MoDL which was reduced to $T=5$ to fit in 12 GB GPU memory.
For MoDL, only the time for the second part of the two-step training has been measured.
The results are presented in \figref{fig:timing}.
In general, the computation time of the BART and TensorFlow implementations are comparable, however, TensorFlow performs better on the older GeForce GTX TITAN X GPU.  BART's implementation of VarNet is distributed to two GPUs by stacking two versions of the network each associated to the respective GPU along the batch dimension (c.f. Supporting Information \figref{supfig:multi_gpu}).
The overhead due to multi-GPU synchronization is minimal resulting in a training time reduced by 47\% to 49\% depending on the GPU.
Batch-normalization used by MoDL requires inter-batch synchronization such that only the data-consistency blocks are distributed to multiple GPUs reducing the benefit of multiple GPUs.  %The network block of MoDL is not distributed to multiple GPUs since the batch-normalization makes slices in one batch depend on each other.
%On the older GPUs MoDL still benefits from the parallelization of the data-consistency block.
%VarNet is fully parallelized over the two GPUs such that training time is almost halved by doubling the number of GPUs. 
The inference time was measured on the respective evaluation datasets.
To reduce the bias due to different pre-processing procedures on the CPU in the respective implementations, we also measured the total execution time of GPU kernels using NVIDIA Nsight Systems.
In general, the BART implementations achieve similar performance to the TensorFlow implementations.

We measured the peak GPU memory allocation during training and inference for the respective implementations of MoDL and VarNet and present the results in \tblref{tbl:memory}.
Since allocating GPU memory is expensive, both, BART and TensorFlow, use a memory cache to reuse allocated memory.
While BART allocates memory on demand, TensorFlow pre-allocates larger memory blocks such that the peak memory allocation exceeds the actually required memory.
Thus, we also state the memory allocation before reaching the peak allocation which serves as a loose lower bound.      The memory needed for training and inference is overall similar across implementations.
The TensorFlow implementation of VarNet computes the gradient steps in the data-consistency block without removing frequency oversampling which is a possible reason for the higher memory requirement.
Results on the TITAN Xp are based on an old version of BART since the GPU broke during the revision.    %\begin{table}

\begin{table}
    \centering
    \caption{                  GPU-Memory (in GB) used by BART and TensorFlow to train/infer MoDL and VarNet on different GPUs (full names in text).
        In parentheses, we provide for BART the memory if cuDNN is used and for TensorFlow the memory used before the peak-allocation is reached, which serves as a loose lower bound.                           \label{tbl:memory}
    }\vspace*{\baselineskip}

    %\resizebox{\linewidth}{!}{
    \begin{tabular}{rcccc}
        \toprule
        &\multicolumn{2}{c}{\textbf{Training}} & \multicolumn{2}{c}{\textbf{Inference}}\\
        \midrule
        MoDL & BART & TensorFlow & BART & TensorFlow \\
        \hline
        A100       & 8.9 (9.6) & 10.2 (5.8) & 1.8 (2.2) & 1.9 (1.4) \\
        V100 & 8.6 (8.9) &  9.6 (2.9) & 1.6 (1.7) & 1.4 (1.1) \\
        TITAN Xp            &      11.8 (12.6)             &  9.2 (4.9) &      1.7 (1.8)             & 4.0 (0.7)  \\
        TITAN X         & 8.3 (9.0) &  9.2 (4.9) & 1.1 (1.1) & 4.0 (0.6)  \\
        \midrule
        VarNet & BART & TensorFlow & BART & TensorFlow \\
        \midrule
        A100       & 6.6 (6.8) & 18.7 (10.2) & 1.6 (2.0) & 1.9 (1.7) \\
        V100& 6.2 (6.3) & 18.2  (9.6) & 1.3 (1.6) & 1.4 (1.1) \\
        TITAN Xp            &      6.3 (6.3)             & 12.4  (9.2) &      1.2 (1.4)             & 1.1 (0.8) \\
        TITAN X & 5.8 (5.8) & 12.4  (9.2) & 1.0 (1.0) & 1.0 (0.5) \\
        \bottomrule
    \end{tabular}
    %}
\end{table}

\subsection{Image Reconstruction using a Learned Prior}
\begin{figure}
    \centering
    \includegraphics[width=.7\linewidth]{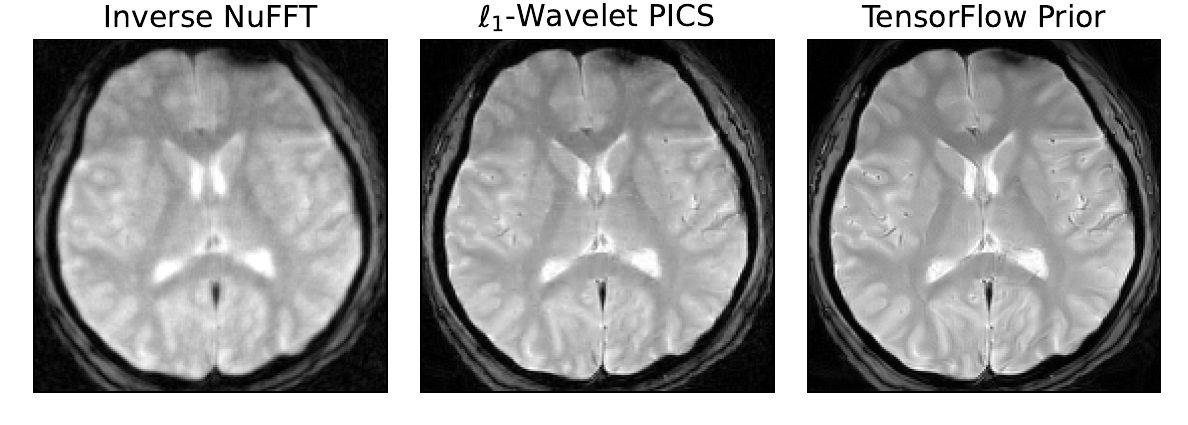}
    \caption{Brain images reconstructed from 60 radial k-space spokes via a coil-combined inverse nuFFT, an $\ell_1$-Wavelet regularized PICS reconstructions, and a PICS reconstruction using a learned log-likelihood prior (left to right).}
    \label{fig:fig-tf}
\end{figure}

In \figref{fig:fig-tf}, we present the reconstruction based on a learned prior. The prior was retrained on the brain dataset described in \cite{Luo_Magn.Reson.Med._2020}. Data for reconstruction was acquired on a Siemens Skyra 3T scanner (Siemens Healthcare GmbH, Erlangen, Germany).
For reconstruction, we used 60 spokes of a radial Flash sequence (TR=770ms, TE=16ms, FA=18$^\circ$).
The coil sensitivity maps were estimated using ESPIRiT and gradient delays were corrected based on RING \cite{Rosenzweig_Magn.Reson.Med._2018a}.
We compare a reconstruction using the inverse nuFFT, an $\ell_1$-Wavelet regularized PICS reconstruction and a reconstruction using the learned log-likelihood prior (c.f. \eqref{eq:reg_pixelcnn}).
The learned log-likelihood prior results in an improved reconstruction compared to the classical $\ell_1$-Wavelet regularized reconstruction.

\section{Discussion}

In this work, we describe a framework for deep learning that
was included in the BART toolbox. The framework is based on
an extension of the existing non-linear operator framework in BART
that provides automatic differentiation and directly integrates
BART's existing MRI-specific operators such as multidimensional
FFT, nuFFT, and SENSE operators and complements it with many
operators commonly used to construct neural networks.
A sophisticated framework for constructing complex neural networks
was added. We also implemented various optimizing techniques and
achieve computational performance similar to other deep-learning frameworks.
Distributing computation to multiple GPUs can reduce computation time further.
Finally, we added new optimization algorithms such as stochastic
gradient descent, iPALM, and Adam, which are popular for training
neural networks.
To demonstrate practicality of the framework, we implemented and trained the Variational Network and MoDL in BART.
Our implementation achieves similar performance in terms of reconstruction quality and training time compared to the original implementations based on TensorFlow.
Further, BART's generic formulation of the SENSE model including the non-Cartesian and soft-SENSE formulation together with a flexible parametrization of the training procedure in terms of training algorithm, loss functions, and training target (coil combined reconstruction, RSS reconstruction or coil images) enables direct use of VarNet and MoDL for many applications.

State-of-the-art deep-learning-based MR image reconstruction algorithms combine two fields of research, i.e. the field of machine learning and the
field of classical MRI reconstruction methods.
For both fields, mature software frameworks/toolboxes already exist.
Hence, various approaches exist to develop algorithms combining both fields, the
two extreme cases are
1.) MRI specific operations can be re-implemented in deep-learning frameworks
and 2.) neural networks can be re-implemented in MRI frameworks.
Both approaches have different advantages and disadvantages.
Deep learning frameworks such as TensorFlow or PyTorch are driven by large communities 
and recent developments in the field of deep learning are quickly integrated.
Moreover, many tutorials based on standard frameworks exist and simple scripting based on Python reduces the barrier to entry.
The frameworks are designed for large scale datasets and support most recent hardware as well as direct integration into cloud solutions of
various providers.
However, all these features come at a price:
Current deep learning libraries use many external libraries with complex dependencies.
Updating some libraries in the backend or the framework itself might produce version conflicts which are hard to resolve.
Long-term reproducibility of research results is difficult to achieve in this environment.
One solution is to freeze the environment in a software container
which contains the specific software versions that are known to work.
In this way, containers can facilitate the reproduction of results and the
translation of working setups to clinical pipelines \cite{Hansen_Magn.Reson.Med._2013, Xue__2019}.
While freezing setups is a legitimate approach for production environments or reproducing results,
it is not a sustainable solution for long term research and development,
where new developments need to build on top of existing code.

On the other side, BART is designed for rapid prototyping, reproducible research and clinical translation.
It depends only on a few external libraries such as FFTW, BLAS implementations or --- if compiled with GPU support --- CUDA, making 
it simple to integrate into different software environments.
Where standard deep learning frameworks benefit from large community support integrating new deep learning features,
BART benefits from years of research on MR image reconstruction. For example, a crucial part of most multi-coil
reconstruction networks is the estimation of the coil sensitivity maps in a preprocessing step.
Since BART implements several calibration methods such as ESPIRiT or NLINV, a full reconstruction
pipeline based on Variational Network or MoDL can be implemented completely in BART.
Advanced concepts from MRI implemented in BART can be directly used in these
machine learning methods. For example, our data consistency modules are implemented
using a generalized SENSE model supporting multiple sets of coil sensitivity maps, non-Cartesian trajectories  and higher dimensional extensions which have been shown beneficial for dynamic MRI \cite{Kuestner_Sci.Rep._2020, Terpstra__2022} .
Concerning performance of training neural networks, we have demonstrated that BART
can compete with the TensorFlow implementation of MoDL and VarNet.  We hope that the deep integration of MRI-specific operators will be appreciated by researchers from other groups such that they will contribute to this open-source deep-learning framework.
Further, we plan to reduce the entry barrier for possible users by extending the TensorFlow wrapper such that it can be used for defining denoising networks which can then be combined with BART's data-consistency modules in the \texttt{reconet} command.  BART development makes uses of a continuous integration framework that uses
automatic testing. Based on this, we aim for long-term reproducibility of
published results even with new BART versions.
Finally, BART is already widely used in the community of MRI research and also
used for clinical research as part of automatic reconstruction frameworks
such as Gadgetron \cite{Hansen_Magn.Reson.Med._2013, Diakite__2018} or Yarra \cite{Block__2016}.
Thus, we believe that the integration of neural networks into BART will also
facilitate research and clinical translation of deep learning methods for image reconstruction.

\section{Conclusion}

By integrating a complete set of tools for training and using neural networks
into BART, we provide a general framework for research in image reconstruction 
that combines state-of-the-art methods for image reconstruction with
deep-learning-based methods. The implementation of two
recent deep-learning-based methods in BART demonstrates
similar performance as their original TensorFlow-based implementations.

\section*{Conflict of Interest}
The authors declare no competing interests.

\section*{Open Research}
\subsection*{Data Availability Statement}
In the spirit of reproducible research, code to reproduce the experiments are available on \url{https://github.com/mrirecon/deep-deep-learning-with-bart}.
BART itself is available on \url{https://github.com/mrirecon/bart}.
The data that support the findings of this study are available from the publications of the Variational Network \cite{Hammernik_Magn.Reson.Med._2017} and MoDL \cite{Aggarwal_IEEETrans.Med.Imag._2019}.
(Converted) data are available at \doi{10.5281/zenodo.6482960} (VarNet) and \doi{10.5281/zenodo.6481291} (MoDL).

\section*{Acknowledgements}
\printfunding
We gratefully acknowledge the support of the NVIDIA Corporation with the donation of one NVIDIA TITAN Xp GPU for this research.

\clearpage
\appendix
\section*{Supporting Information}

\begin{figure*}[h]
	\includegraphics[width=\linewidth]{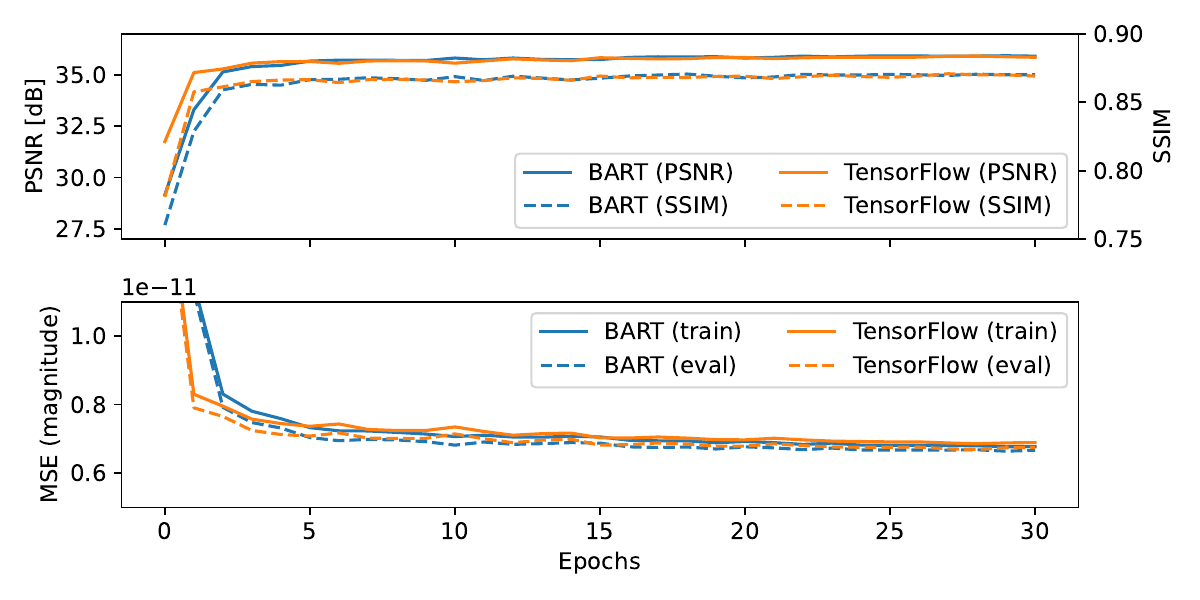}
	\caption{\textbf{Top}: Mean PSNR and SSIM evaluated on the 100 slices of the VarNet evaluation dataset after each training epoch. \textbf{Bottom}: Similarly, MSE of magnitude images evaluated on the training dataset (300 slices) and evaluation dataset (100 slices). Both, the BART and the TensorFlow implementation, show a similar convergence behavior. We assume the slight difference of both metrics in the early epochs result from a different initialization of the weights in both implementations.}
	\label{supfig:train_history}
\end{figure*}

\begin{figure*}[h]
	\includegraphics[width=\linewidth]{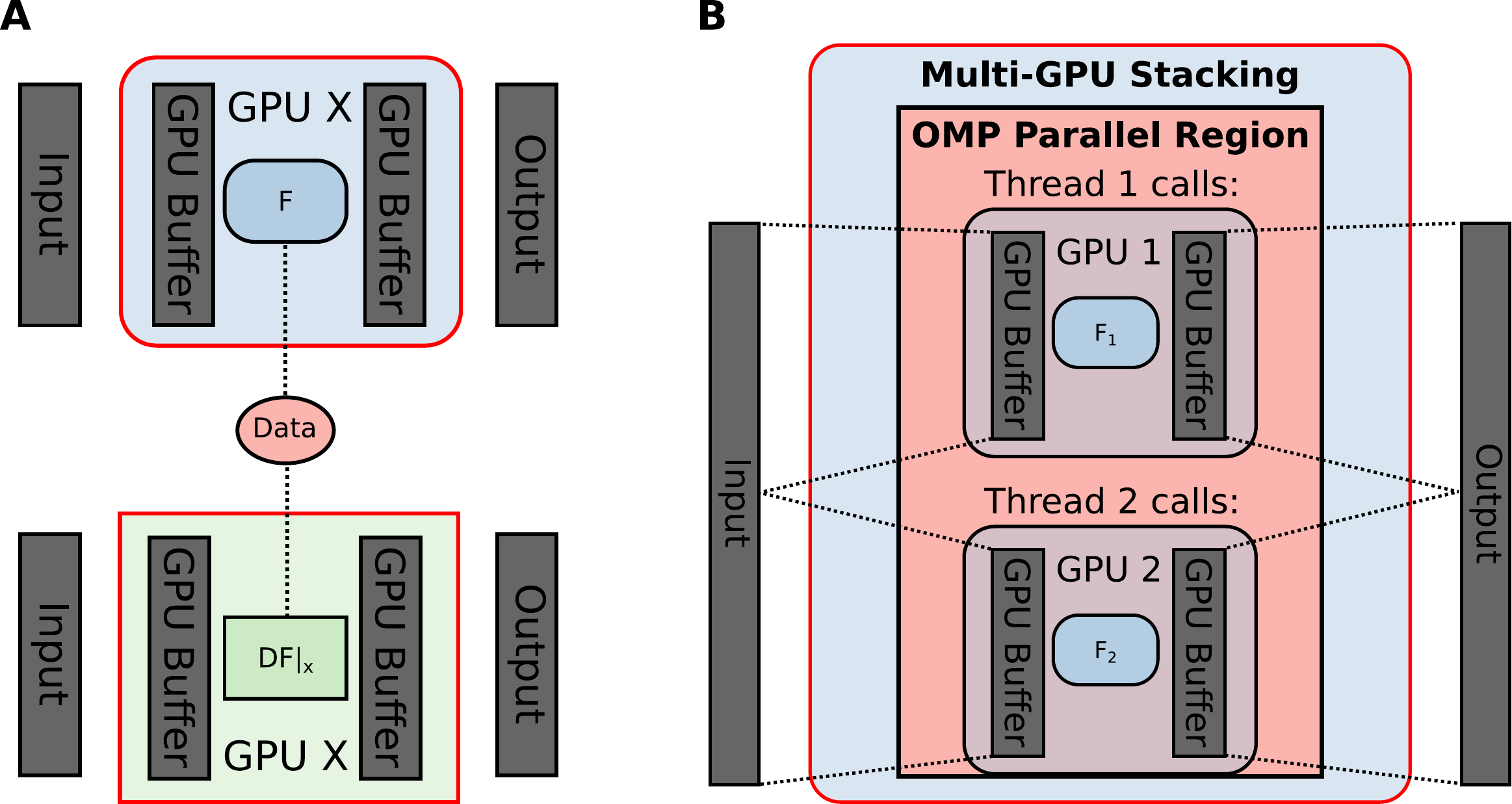}
	\caption{\textbf{A}: An \texttt{nlop}-container assigning the \texttt{nlop} $F$ to GPU X. When the container is called, it changes the CUDA context to the GPU X. CUDA events are used to (asynchronously with respect to the CPU) synchronize the new CUDA context with the old one. The input data is copied to a GPU buffer. Now, the container calls $F$ in the new CUDA context such that all data shared with the derivative is allocated on GPU X. Finally, the output is copied to the output array, before the CUDA context is switched back to the original one. The input and output arrays can be located on the CPU or an arbitrary GPU. \textbf{B}: Multi-GPU stacking of \texttt{nlop}s $F_1$ and $F_2$. The \texttt{nlops} $F_1$ and $F_2$ are assigned to different GPUs and are called in parallel by the corresponding OMP-threads. Both GPU wrappers are synced with the CUDA context active before entering the OMP parallel region. Before leaving the OMP region, the respective CPU threads are synchronized with the CUDA context such that the CUDA context active after leaving the OMP region can assume that all data is written to the output.}
	\label{supfig:multi_gpu}
\end{figure*}

\end{document}